\documentclass[]{article}

    \PassOptionsToPackage{numbers, compress}{natbib}

\usepackage[preprint]{neurips_2021}

\usepackage[utf8]{inputenc} 
\usepackage[T1]{fontenc}    
\usepackage{hyperref}       
\usepackage{url}            
\usepackage{booktabs}       
\usepackage{amsfonts}       
\usepackage{nicefrac}       
\usepackage{microtype}      
\usepackage{xcolor}         
\usepackage{graphicx}
\usepackage{booktabs} 

\title{Rethinking binary hyperparameters for deep transfer learning for image classification}

\author{%
  Jo Plested\\
  Department of Computer Science\\
  University of New South Wales\\
  Northcott Dr, Campbell ACT 2612 \\
  \texttt{j.plested@unsw.edu.au} \\
  \And
  Xuyang Shen\\
  School of Computer Science\\
  Australian National University \\
  \texttt{xuyang.shen@anu.edu.au} \\
  \And
  Tom Gedeon\\
  School of Computer Science\\
  Australian National University \\
  \texttt{tom.gedeon@anu.edu.au} \\
}

\begin{document}

\maketitle

\begin{abstract}
 The current standard for a variety of computer vision tasks using
smaller numbers of labelled training examples is to fine-tune from
weights pre-trained on a large image classification dataset such as
ImageNet. The application of transfer learning and transfer learning
methods tends to be rigidly binary. A model is either pre-trained
or not pre-trained. Pre-training a model either increases performance or decreases it, the latter being defined as negative transfer. Application of L2-SP regularisation that decays the weights towards their pre-trained values is either applied or all weights are decayed towards 0. This paper re-examines these assumptions. Our recommendations are based on extensive empirical evaluation that demonstrate the application of a non-binary approach to achieve optimal results. (1) Achieving best performance on each individual dataset requires careful adjustment of various transfer learning hyperparameters not usually considered, including number of layers to transfer, different learning rates for different layers and different combinations of L2SP and L2 regularization. (2) Best practice can be achieved using a number of measures of how well the pre-trained weights fit the target dataset to guide optimal hyperparameters. We present methods for non-binary transfer learning
including combining L2SP and L2 regularization and performing non-traditional fine-tuning hyperparameter searches. Finally we suggest heuristics for determining the optimal transfer learning hyperparameters. The benefits of using a non-binary approach are supported by final results that come close to or exceed state of the art performance on a variety of tasks that have traditionally been more difficult for transfer learning.
\end{abstract}

\section{Introduction}

Convolutional neural networks (CNNs) have achieved many successes
in image classification in recent years \cite{krizhevsky2012imagenet,girshick2014rich,li2020deep,masi2018deep,mazurowski2019deep}.
It has been consistently demonstrated that CNNs work best when there
is abundant labelled data available for the task and very deep models
can be trained \cite{ngiam2018domain,mahajan2018exploring,kolesnikov2019big}.
However, there are many real world scenarios where the large amounts
of training data required to obtain the best performance cannot be
met or are prohibitively expensive. Transfer learning has been shown
to improve performance in a wide variety of computer vision tasks,
particularly when the source and target tasks are closely related
and the target task is small \cite{ngiam2018domain,mahajan2018exploring,cui2018large,ge2017borrowing,sabatelli2018deep,ng2015deep,li2018explicit,li2019delta,wan2019towards}. It has become
standard practice to pre-train on Imagenet 1K for many different tasks
where the available labeled datasets are orders of magnitude smaller
than Imagenet 1K \cite{li2018explicit,li2019delta,wan2019towards,masi2018deep,li2020deep,ng2015deep,mazurowski2019deep,mormont2018comparison,girshick2014rich}. Several papers published in
recent years have questioned this established paradigm. They have
shown that when the target dataset is very different from Imagenet
1K and a reasonable amount of data is available, training from scratch
can match or even out perform pre-trained and fine-tuned models \cite{he2018rethinking,shen2017learning,shen2017dsod,zhu2019scratchdet,mahajan2018exploring}.

The standard
transfer learning strategy is to transfer all layers apart from the
final classification layer, and either use a single initial learning rate and other hyperparameters for fine-tuning all layers, or freeze some layers. Given that lower layers in a deep neural network are known to be more
general and higher layers more specialised \cite{yosinski2014transferable},
we argue that the binary way of approaching transfer learning and
fine-tuning is counter intuitive. There is no intuitive reason to
think that all layers should be treated the same when fine-tuning,
or that pre-trained weights from all layers will be applicable to
the new task. If transferring all layers results in negative transfer,
could transferring some number of lower more general layers improve
performance? If using an L2SP weight decay on all transferred layers
for regularisation decreases performance over decaying towards 0,
might applying the L2SP regularisation to some number of lower layers
that are more applicable to the target dataset result in improved
performance? 

We performed extensive experiments across four different datasets
to: 
\begin{itemize}
\item re-examine the assumptions that transfer learning hyperparameters should be binary, and 
\item find the optimal settings for number of layers to transfer, initial
learning rates for different layers, and number of layers to apply
L2SP regularisation vs decaying towards 0. 
\end{itemize}
We developed methods for non-binary transfer learning including combining
L2SP and L2 regularization and performing non-traditional fine-tuning
hyperparameter searches. We show that the optimal settings result
in significantly better performance than binary settings for all datasets except the most closely related. Finally we suggest heuristics for determining the optimal transfer learning hyperparameters.

\section{Related work}

The standard
transfer learning strategy is to transfer all layers apart from the
final classification layer, then use a search strategy to find the
best single initial learning rate and other hyperparameters. Several
studies include extensive hyperparameter searches over learning rate
and weight decay \cite{kornblith2019better,mahajan2018exploring},
momentum \cite{li2020rethinking}, and L2SP \cite{li2018explicit}.
This commonly used strategy originates from various works showing
that performance on the target task increases as the number of layers
transferred increases \cite{yosinski2014transferable,chu2016best,agrawal2014analyzing,azizpour2015factors}.
All these works were completed prior to advances in residual
networks \cite{he2016deep}, and other advances to improve learning for deep CNN architectures, and searches for optimal combinations of learning rates and number of layers to transfer were not performed.
Additionally, in two of the works layers that were not transferred
were discarded completely rather than reinitialising and training
them from scratch. This resulted in smaller models with less layers
transferred and a strong bias towards better results when more layers
were transferred \cite{agrawal2014analyzing,azizpour2015factors}.
Further studies have shown the combination of a lower learning rate
and fewer layers transferred may be optimal \cite{plested2019analysis}.
However, again modern residual networks were not used and only very
similar source and target datasets were selected.

It has been shown that the similarity between source and target datasets has a strong impact on performance for transfer learning:
\begin{enumerate}
\item More closely related datasets can be better than more source data
for pre-training \cite{ngiam2018domain,mahajan2018exploring,cui2018large,ge2017borrowing,sabatelli2018deep}.
\item A multi-step pre-training process where the interim dataset is smaller
and more closely related to the target dataset can outperform a single
step pre-training process when originating from a very different, large
source dataset \cite{ng2015deep}. 
\item Self-supervised pre-training on a closely related source dataset can
be better than supervised training on a less closely related dataset
\cite{zoph2020rethinking}. 
\item L2SP regularization, where the weights are decayed towards their pre-trained
values rather than 0 during fine-tuning, improves performance when
the source and target dataset are closely related, but hinders it
when they are less related \cite{li2018explicit,li2019delta,wan2019towards}
\item Momentum should be lower for more closely related source and target datasets  \cite{li2020rethinking}.
\end{enumerate}

These five factors demonstrate the importance of the relationship
between the source and target dataset in transfer learning. 

\section{Methodology}

\subsection{Datasets}

We performed evaluations on four different small target datasets,
each being less than 10,000 training images. We chose one that is
very similar to Imagenet 1K that transfer learning and sub methods
have traditionally performed best on. This is used as a baseline for
comparison to show that the default methods that perform badly on
the other datasets chosen do perform well on this one. For each of
the other target datasets it has been shown that traditional transfer
learning strategies: 
\begin{itemize}
\item do not perform well on them and/or 
\item they are very different to the source dataset used for pre-training. 
\end{itemize}
We used the standard available train, validation and test splits for the three datasets for which they were available. For Caltech256-30 we used the first 30 items from each class for the train split, the next 20 for the validation split and the remainder for the test split.

\subsubsection{Source dataset}

We used Imagenet 1K as the source dataset as it is most commonly used
source dataset and therefore the most suitable to demonstrate our approach.

\subsubsection{Target datasets}

These are the final datasets that we transferred the models to
and measured performance on. We used a standard 299 $\times$ 299
image size for all datasets.
\paragraph{Most similar to Imagenet 1K}
We chose Caltech 256-30 (Caltech) \cite{griffin2007caltech} as the most similar to Imagenet. It contains 256 different general subordinate and superordinate classes. As our focus is on small target datasets, we chose the smallest commonly used split with 30 examples per class.  

\paragraph{Fine-grained}

Fine-grained object classification datasets contain subordinate classes
from one particular superordinate class. We chose two where standard
transfer learning has performed badly \cite{guo2019spottune,kornblith2019better}.
\begin{itemize}
\item Stanford Cars (Cars): Contains 196 different makes and models of cars
with 8,144 training examples and 8,041 test examples \cite{KrauseStarkDengFei-Fei_3DRR2013}. 
\item FGVC Aircraft (Aircraft): Contains 100 different makes and models
of aircraft with 6,667 training examples and 3,333 test examples \cite{maji13fine-grained}.
\end{itemize}

\paragraph{Very different to Imagenet 1K}

We evaluated performance on another dataset that is intuitively very
different to Imagenet 1K as it consists of images depicting adjectives
instead of nouns. 
\begin{itemize}
\item Describable Textures (DTD) \cite{cimpoi2014describing}: consists
of 3,760 training examples of texture images jointly annotated with
47 attributes. The texture images are collected \textquotedblleft in
the wild\textquotedblright{} by searching for texture adjectives on
the web. 
\end{itemize}

\subsection{Model  \label{model}}
Inception v4 \cite{szegedy2017inception} was selected for evaluation
as it has been shown to have state of the art performance for transferring from Imagenet 1K to a broad range of target tasks \cite{kornblith2019better}.
Our code is adapted from \cite{rw2019timm} and we used the publicly available pre-trained Inception v4 model. Our code is available at \url{https://anonymous.4open.science/r/Non-binary-deep-transfer-learning-for-image-classification-872D}.
We did not experiment with pre-training settings, for example removing
regularization settings as suggested by \cite{kornblith2019better},
as it was beyond the scope of this work and the capacity of our compute resources.

\subsection{Evaluation metric}

We used top-1 accuracy for all results for easiest comparison with
existing results on the chosen datasets. Graphs showing ablation studies
with different hyperparameters show results on one run. Final reported
results are averages and standard deviations over four runs. For all
experiments we used single crop and for our final comparison to state
of the art for each dataset we used an ensemble of all four runs and 10-crop. 

\subsection{Compute resources \label{compute}}

The experiments were run on compute nodes containing four Nvidia V100 GPUs and two 24-core Intel Xeon Scalable 'Cascade Lake' processors. 

Each individual experiment in the ablation studies ran on one GPU and 12 cores from one processor. Running time varied between approximately six hours for DTD to one day for Caltech.

Each final experiment was run on a full compute node with four GPUs and two 24-core processors. Runnng time varied between approximately one day for DTD and Aircraft to two days for Caltech.

\section{Results}

\subsection{Assessing the suitability of the fixed features}

We first examined performance using the pre-trained Inception v4 model as a fixed
feature extractor for comparison and insight into the suitability
of the pre-trained features. We trained a simple
fully connected neural network classification layer for each dataset
and compared it to training the full model from random initialization.
The comparisons for all datasets are shown in Table \ref{domain measures}.

The class normalized Fisher score on target datasets using the pre-trained but not fine-tuned weights, was used as a measure of how well the pre-trained weights separate the classes in the target dataset \cite{fukenaga1990introduction}. A larger normalized Fisher score shows more separation between classes in the feature space.

\[
F\left(\mathbf{W}\right)=Tr\left\{ s_{w}^{-1}s_{B}\right\} /N_{c}
\]

where $s_{w}^{-1}$ is the inverse of the within class covariance,
$s_{B}$ is the between class covariance and $N_{c}$ is the number
of classes for the target dataset being measured. See \cite{fukenaga1990introduction} for further details.

We also calculated domain similarity between the fixed features of
the source and target domain using the earth mover distance (EMD)
\cite{rubner2000earth,peleg1989unified} and applying the procedure defined in \cite{cui2018large}.

The domain similarity calculated using the EMD has been shown to correlate well with the improvement in performance on the target task from pre-training with a particular source dataset \cite{cui2018large}.

\begin{table}[th]

\caption{Domain measures}
\label{domain measures}
\centering
\begin{tabular}{p{1.5cm}p{1.8cm}p{1.8cm}p{1.8cm}p{2cm}p{2.5cm}}
\hline 
 & {\small{}trained from random initialization} & {\small{}fixed features classification} & {\small{}EMD domain similarity} & normalised Fisher score & {\small{}state of the art}\tabularnewline
\hline 
{\small{}Caltech} & 67.2 & 83.4  & 0.568 & 1.35 & 84.9 \cite{li2019delta}\tabularnewline
{\small{}Cars} & 92.7 & 64.2 & 0.536 & 1.18 & 96.0 (@448) \cite{ridnik2020tresnet}\tabularnewline
{\small{}Aircraft} & 88.8 & 59.9 & 0.557 & 0.83 & 93.9 (@448) \cite{zhuang2020learning}\tabularnewline
{\small{}DTD} & 66.8 & 74.6 & 0.540 & 3.47 & 78.9 \cite{chen2020simple}\tabularnewline
\hline 
\end{tabular}

\end{table}

\subsection{Default settings}

\begin{figure}[th]

\begin{centering}
\includegraphics[scale=0.22]{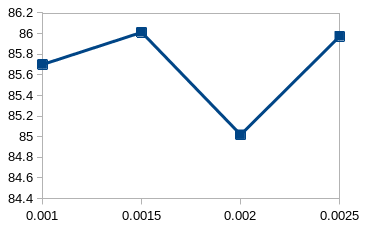}\includegraphics[scale=0.23]{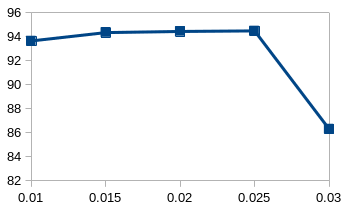}\includegraphics[scale=0.22]{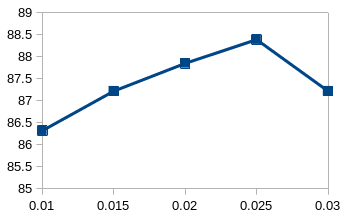}\includegraphics[scale=0.22]{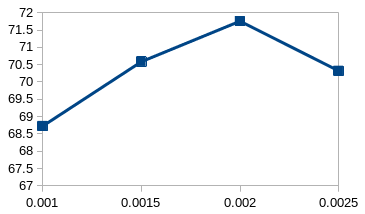}
\caption{Optimal learning rates for each dataset. Left to right Caltech, Cars, Aircraft, DTD}
\label{single learning rates}
\par\end{centering}
\end{figure}

For the first experiment we transferred all but the final classification
layer and trained all layers at the same learning rate as per standard
transfer learning practice. We performed a search using the validation
set to find the optimal single fine-tuning learning rate for each
dataset. The results in Figure \ref{single learning rates} show the accuracy for each learning rate for each dataset.

The optimal single learning rate for Caltech, the most similar dataset
to ImageNet, is an order of magnitude lower than the optimal learning
rate for the fine-grained classification tasks Stanford Cars and Aircraft.
This shows that the optimal weights for Caltech are much closer to
the pre-trained values. The surprising result was that the optimal
learning rate for DTD was very similar to Caltech and also an order
of magnitude lower than Stanford Cars and Aircraft. Final results on the test set for each dataset are shown in Table
\ref{final default}.

\begin{table}[th]

\caption{Final results default settings}
\label{final default}
\centering
\begin{tabular}{ccccc}
\hline 
\centering
 & Caltech & Stanford Cars & Aircraft & DTD\tabularnewline
\hline 
Learning rate & 0.0025  & 0.025 & 0.025 & 0.002 \tabularnewline
Accuracy & 83.69$\pm$0.0784 & 94.59$\pm$0.110 & 93.78$\pm$0.137 & 77.30 $\pm$0.726\tabularnewline
\hline 
\end{tabular}
\end{table}

\subsection{Optimal learning rate decreases as more layers are reinitialized
and transferring all possible layers often reduces performance on
the target task}

\begin{figure}[th]

\begin{centering}
\label{layers vs lr}
\includegraphics[scale=0.27]{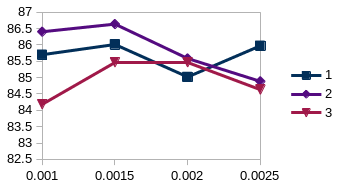}\includegraphics[scale=0.27]{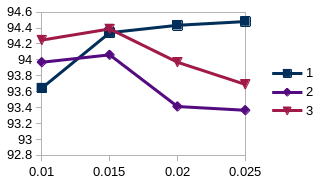}\includegraphics[scale=0.27]{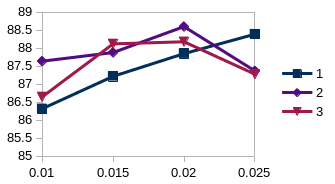}\includegraphics[scale=0.27]{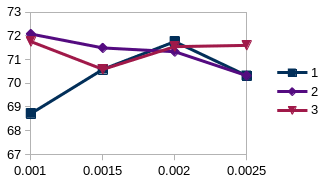}
\caption{Number of layers reinitialized versus learning rate. Left to right Caltech, Cars, Aircraft, DTD }
\par\end{centering}
\end{figure}

To the best of our knowledge the combination of learning rate and
number of layers reinitialized when performing transfer learning has
not been examined in modern residual models. To examine the relationship
between learning rate and number of layers reinitialized, we searched
for the optimal learning rate for each of 1-3 blocks of layers reinitialized
across each of the four datasets. One layer is the default with the
final classification layer only being reinitialized. Two and three
involve reinitializing an additional one or two Inception C blocks
of layers respectively as well as the final layer. For consistency
we refer to both the final classification layer and the Inception
C blocks as layers. Figure \ref{layers vs lr} shows the performance of fine-tuning
with various combinations of learning rate and layers reinitialized.
The optimal learning rate when reinitializing more than one layer
is always lower than when reinitializing only the final classification
layer. Also the optimal number of layers to reinitialize is more than
one for all datasets except Cars. Table \ref{final layers vs lr} shows the final results
for the optimal learning rate and accuracy for each number of layers
reinitialized. 

\paragraph{Reinitializing more layers has a more significant effect when the
optimal learning rate for reinitializing just one layer is higher}

Caltech and DTD have an order of magnitude lower optimal learning
rates than Cars and Aircraft, but reinitializing more layers for the
former results in a significant increase in accuracy whereas there
is none for the latter. This result initially seems counter intuitive
as Cars and Aircraft are less similar to Imagenet than Caltech and
DTD. However, a lower learning rate tempers the ability of the upper
layers of the model to specialise to the new domain and even very
closely related source and target domains have different final classes
and thus optimal feature spaces. The combination of a lower learning
rate and reinitializing more layers likely allows the models to keep
the closely related lower and middle layers and create new specialist
upper layers.

\begin{table}[th]

\caption{Final optimal learning rate for each number of layers reinitialized}
\label{final layers vs lr}
\centering
\begin{tabular}{p{2cm}p{2.3cm}p{2.3cm}p{2.3cm}p{2.3cm}}
\hline 
 & Caltech & Cars & Aircraft & DTD\tabularnewline
\hline 
1 layer lr & 0.0025 & 0.025 & 0.025 & 0.002\tabularnewline
Accuracy & 83.69$\pm$0.078 & 94.59$\pm$0.110 & 93.78$\pm$0.137 & 77.30 $\pm$0.726\tabularnewline
2 layers lr & 0.015 & 0.015  & 0.02 & 0.001\tabularnewline
Accuracy & 84.31$\pm$0.114 & 94.59$\pm$ 0.205 & 93.56$\pm$0.942 & 78.92 $\pm$0.191\tabularnewline
3 layers lr & 0.015 & 0.015  & 0.02 & 0.0015 \tabularnewline
Accuracy & 83.57$\pm$0.104 & 94.46$\pm$0.348 & 92.96$\pm$1.588 & 78.90$\pm$0.243\tabularnewline
\hline 
\end{tabular}
\end{table}

\subsection{Lower learning rates for lower layers works better when optimal learning
rate is higher}

Conventionally learning rates for different layers are set so that either
all layers are fine-tuned with the same learning rate or some are
frozen. The thinking is that as lower layers are more general and
higher layers more task specific \cite{yosinski2014transferable}
the lower layers do not need to be fine-tuned. Recent work has shown
that fine-tuning tends to work better in most cases \cite{plested2019analysis}.
However, setting different learning rates for different layers is not
generally considered. We examined the effects of applying lower learning
rates to lower layers that are likely to generalise better to the
target task. Figure \ref{low_lr} shows that for the Stanford Cars and
Aircraft where the optimal initial learning rate is higher, setting
lower learning rates for lower layers significantly improves performance
whereas for Caltech and DTD it does not. 

\begin{figure}[th]

\begin{centering}
\par\end{centering}
\begin{centering}
\includegraphics[scale=0.21]{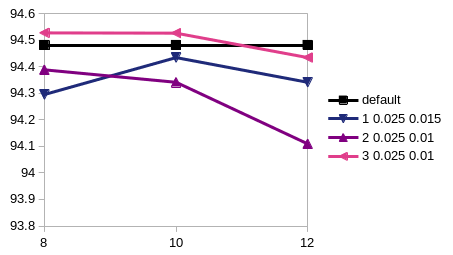}\includegraphics[scale=0.21]{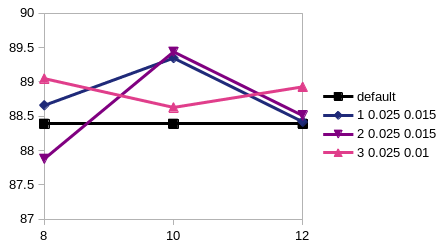}\includegraphics[scale=0.21]{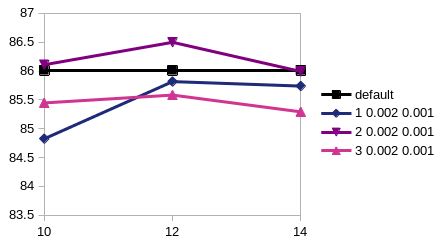}\includegraphics[scale=0.21]{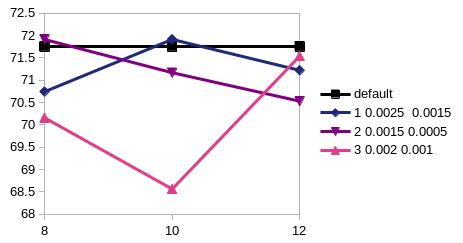}
\caption{Lower learning rates for lower layers. Left to right Caltech, Cars, Aircraft, DTD. Legend shows number of layers reinitialized, high learning rate, low learning rate. X axis is number layers trained starting with the low learning rate}
\par\end{centering}
\end{figure}

\begin{table}[th]

\caption{Optimal learning rates for each number of layers reinitialized with
different learning rates for lower layers. Learning rates are in the format (high layers learning rate, low layers learning rate, number of low layers)
}
\label{low_lr}
\centering
\begin{tabular}{p{2.5cm}p{2.2cm}p{2.2cm}p{2.4cm}p{2.4cm}}
\hline 
 & Caltech & Cars & Aircraft & DTD\tabularnewline
\hline 
Default accuracy  & 83.69$\pm$0.0784 & 94.59$\pm$0.110 & 93.78$\pm$0.137 & 77.30$\pm$0.726\tabularnewline
1 layer lrs & 0.002 0.001 12  & 0.025 0.01 8  & 0.025 0.015 10  & 0.0025 0.0015 10 \tabularnewline
Accuracy & 83.95$\pm$0.196 & 94.86$\pm$0.0460 & 94.32 $\pm$0.144 & 77.753$\pm$0.456\tabularnewline
2 layers lrs & 0.002 0.001 12  & 0.025 0.01 8  & 0.025 0.015 10  & 0.0015 0.001 14 \tabularnewline
Accuracy & 84.15$\pm$0.553 & 94.78 $\pm$0.132 & 94.17 $\pm$0.311 & 78.59$\pm$0.127\tabularnewline
3 layers lrs & 0.002 0.001 12  & 0.025 0.01 10  & 0.01 0.025 0.01 8  & 0.0015 0.001 12 \tabularnewline
Accuracy & 83.46$\pm$0.143 & 94.83 $\pm$0.0832. & 94.50 $\pm$0.192 & 78.84 $\pm$0.464\tabularnewline
\hline 
\end{tabular}
\end{table}

\subsection{L2SP}

The L2SP regularizer decays pre-trained weights towards their pre-trained
values rather than towards zero during fine-tuning. In the standard
transfer learning paradigm with only the final classification reinitialized,
when the L2SP regularizer is applied the final classification layer
only is decayed towards 0.

The values $\alpha$ and $\beta$ are tuneable hyperparameters to control
the amount of regularization applied to the pre-trained and randomly
initialized layers respectively.

The original experiments showing the effectiveness of the L2SP regularizer
\cite{li2018explicit} were done on target datasets that are extremely
similar to the source datasets used. They showed that a high level of
regularization decaying towards the pre-trained weights is beneficial on these datasets.
It has since been shown that the L2SP regularizer can result in minimal
improvement or even worse performance when the source and target datasets
are less related \cite{li2020rethinking,wan2019towards}. 

\begin{figure}[th]

\begin{centering}
\includegraphics[scale=0.20]{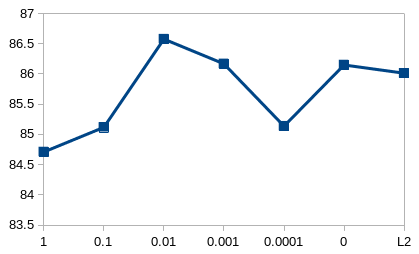}\includegraphics[scale=0.20]{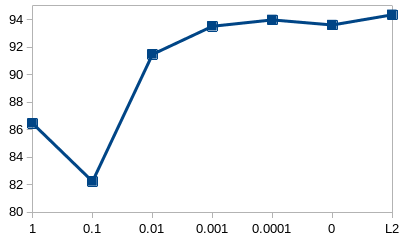}
\includegraphics[scale=0.20]{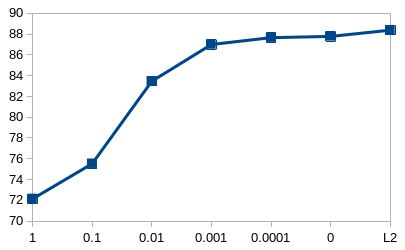}\includegraphics[scale=0.20]{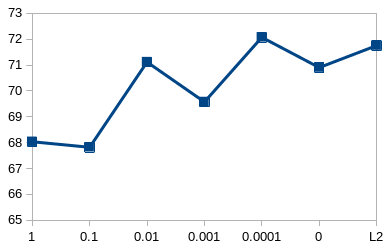}
\caption{L2SP with default settings. Left to right Caltech, Cars, Aircraft, DTD}
\label{L2SP default}
\par\end{centering}
\end{figure}

Our results shown in Figure \ref{L2SP default} align with the original paper
\cite{li2018explicit} for the dataset Caltech showing that standard
L2SP regularization does result in an improvement in performance over
L2 regularization. Our results also align with \cite{li2020rethinking,wan2019towards}
in showing that for the datasets we have chosen to be different from
Imagenet, and known to be more difficult to transfer to, L2SP regularization
performs worse than L2 regularization. In general the lower the setting
for alpha (the L2SP regularization hyperparameter) the better the
performance.

\begin{figure}[th]

\begin{centering}
\includegraphics[scale=0.2]{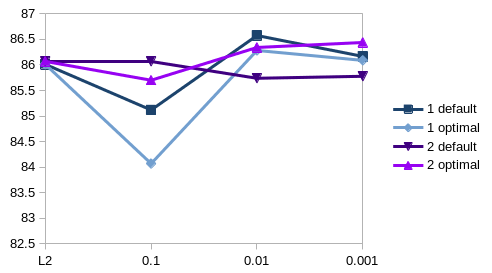}\includegraphics[scale=0.2]{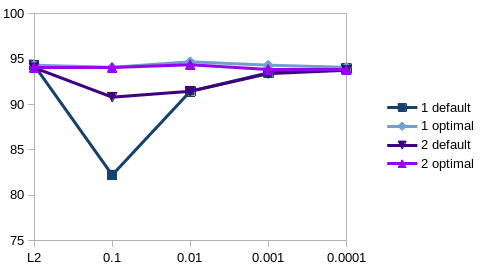}\includegraphics[scale=0.2]{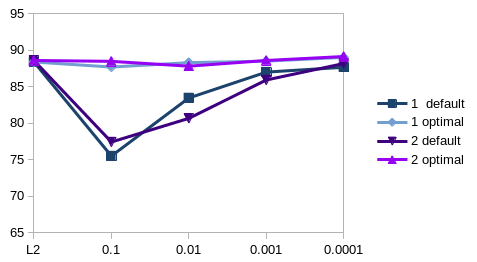}\includegraphics[scale=0.2]{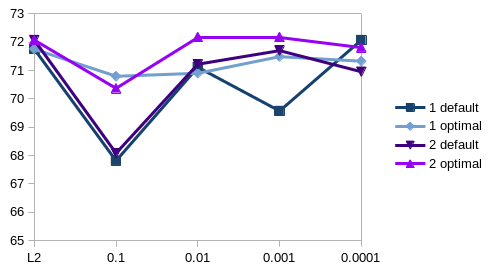}
\caption{Best default L2SP optimal L2SP settings. Left to right Caltech, Cars, Aircraft, DTD}
\label{L2SP optimal figure}
\par\end{centering}
\end{figure}

We relaxed the binary assumption that L2SP regularization must be
applied to all pre-trained layers. We used L2SP regularization for
lower layers that we expected to be more similar to the source dataset
and L2 regularization for upper layers to allow them to specialise to
the target dataset. We searched for the optimal combinations of L2SP
and L2 weight regularization along with the $\alpha$ and $\beta$
hyperparameters for each dataset. We show the best settings for number
of layers and amount of L2SP regularization in Table \ref{L2SP optimal} and
graph the optimal settings compared to the best default settings in
Figure \ref{L2SP optimal figure}. 

We make the following observations:
\begin{enumerate}
\item A combination of L2SP and L2 regularization is optimal for most settings
of the L2SP regularization hyperparameter ($\alpha$) for all datasets
except Caltech. 
\item When more layers are trained with L2 rather than L2SP regularization
the optimal L2 regularization hyperparameter ($\beta$) is lower as
the squared sum of the weights in these layers will be larger for
the same model. 
\end{enumerate}
Further experiments were performed to search for the combination of
optimal L2SP vs L2 regularization settings with optimal number of
layers transferred and different learning rates for different layers.
These results are also shown in Table  \ref{L2SP optimal}. 

\begin{table}[th]

\caption{Best default and optimal L2SP settings and results}
\label{L2SP optimal}
\centering
\begin{tabular}{p{2.2cm}p{2.3cm}p{2.3cm}p{2.3cm}p{2.3cm}}
\hline 
 & Caltech & Cars & Aircraft & DTD\tabularnewline
\hline 
L2 & 83.694  & 94.59 & 93.78 & 77.30\tabularnewline
Default L2SP 1 new layer $\alpha$, $\beta$ & 0.01 0.01  & 0.0001 0.01  & 0.0001 0.01  & 0.0001 0.01 \tabularnewline
result & 84.52 & 94.42  & 93.29 & 78.32\tabularnewline
\hline 
Optimal L2SP 1 new layer $\alpha$, $\beta$ & 0.01 0.01

L2SP layers all & 0.01 0.001 L2SP layers 10 & 0.0001 0.001 L2SP layers 10  & 0.001 0.001 

L2SP layers 10 \tabularnewline
result & 84.52 & 94.57  & 93.86 & 78.32\tabularnewline
\hline 
Optimal L2SP overall $\alpha$, $\beta$ & 0.01 0.01

L2SP layers all & 0.01 0.001 

L2SP layers 10  & 0.0001 0.001 

L2SP layers 10 & 0.001 0.001 1.0 L2SP layers 14\tabularnewline
result  & 84.52 & 94.65 & 94.22 & 78.90\tabularnewline
\hline 
\end{tabular}
\end{table}

\subsection{Final optimal settings and how to predict them}

\begin{figure}[th]

\begin{centering}
\includegraphics[scale=0.27]{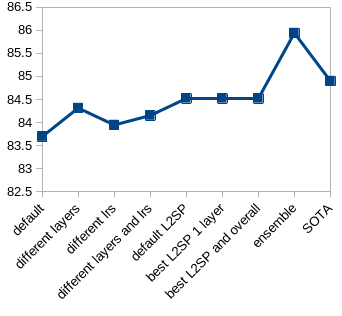}\includegraphics[scale=0.26]{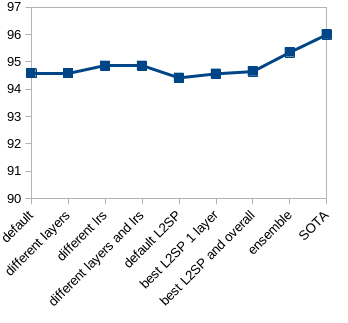}\includegraphics[scale=0.26]{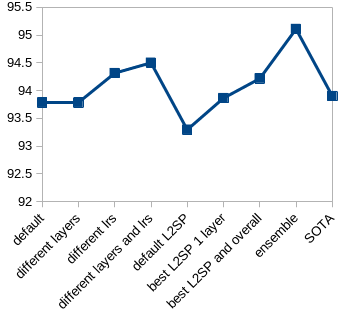}\includegraphics[scale=0.26]{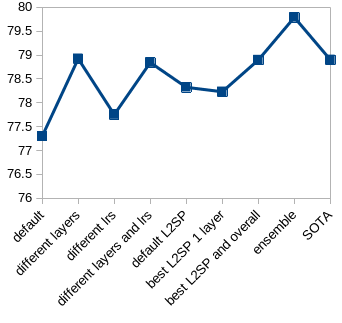}
\caption{Optimal settings versus best default settings. Left to right Caltech, Cars, Aircraft, DTD}
\label{final figure}
\par\end{centering}
\end{figure}

The final results and optimal settings are shown in Table \ref{final} and
Figure \ref{final figure}. Table \ref{predicting} shows a clear distinction between target
datasets that are more similar to the source dataset and for which
the pre-trained weights are better able to separate the classes in
feature space and those that are less similar with pre-trained weights
that fit the target task poorly. The best measure for determining whether the pre-trained weights are well suited is a comparison of the performance achieved with fixed pre-trained features and that achieved through training from random initialization. As this method is computationally intensive a reasonable alternative may be the normalized Fisher Ratio, but as the differences are not as pronounced in all cases it should be further investigated on more datasets to see how reliable it is as a heuristic. The EMD measure of domain similarity is a poor predictor of the suitability of the pre-trained weights.

Targets datasets for which the pre-trained weights are well suited need: 
\begin{itemize}
\item a much lower learning rate for fine-tuning,
\item more than one layer to
be reinitialized from random weights, and
\item some or all layers trained with L2SP regularization. 
\end{itemize}
Target datasets where the pre-trained weights are not well suited need: 
\begin{itemize}
\item a much higher learning rate for fine-tuning with a lower learning
rate for lower layers,
\item only the final classification layer reinitialized, and
\item L2 rather than L2SP regularization. 
\end{itemize}
Using the above best practice, non-binary transfer learning procedures
we achieved state of the art or close to, on three out of the four
datasets. We used publicly available pre-trained weights and no additional
methods for either pre-training or fine-tuning. 

\begin{table}[t!]

\caption{Optimal settings versus best default settings}
\label{final}
\centering
\begin{tabular}{p{2.3cm}p{2cm}p{2.4cm}p{2.4cm}p{2.2cm}}
\hline 
 & Caltech & Cars & Aircraft & DTD\tabularnewline
\hline 
Default settings & 0.0025  & 0.025  & 0.025  & 0.002 \tabularnewline
Default result & 83.69 & 94.59 & 93.78 & 77.30\tabularnewline
Default L2SP & 84.52 & 94.42 & 93.29 & 78.32\tabularnewline
\hline 
Optimal settings & 1 new layer lr 0.0025 L2SP 0.01 0.01  & 1 new layer

high lr 0.025    low lr 0.01 

low layers 8 

no L2SP & 3 new layers

FC lr 0.01

high lr 0.025   
low lr 0.01 

low layers 8

no L2SP & new layers 2 

lr 0.002 

L2SP 0.001 0.001

L2SP layers 14\tabularnewline
Optimal result & 84.52 & 94.86 & 94.50 & 78.90\tabularnewline
\hline 
Ensemble & 85.94 & 95.35 & 95.11 & 79.79 \tabularnewline
State of the art & 84.9 \cite{li2019delta} & 96.0 (@448) \cite{ridnik2020tresnet} & 93.9 (@448) \cite{zhuang2020learning} & 78.9 \cite{chen2020simple}\tabularnewline
\hline 
\end{tabular}
\end{table}

\begin{table}[th]

\caption{Predicting optimal settings based on pre-trained features}
\label{predicting}
\centering
\begin{tabular}{p{3.2cm}cccc}
\hline 
 & Caltech & DTD & Cars & Aircraft\tabularnewline
\hline 
Fisher ratio & 1.35 & 3.47 & 1.18 & 0.83\tabularnewline
EMD similarity & 0.568 & 0.540 & 0.536 & 0.557\tabularnewline
Random initialisation minus fixed features & -16.2 & -7.8 & 28.5 & 28.9\tabularnewline
Optimal learning rate & Low & Low & High & High\tabularnewline
L2SP & Yes & Yes & No & No\tabularnewline
More layers reinitiailized & Yes & Yes & No & No\tabularnewline
Low layers at low learning rate & No & No & Yes & Yes\tabularnewline
\hline 
\end{tabular}

\end{table}

\section{Discussion \label{subsec:discussion}} 
Traditional binary assumptions about transfer learning hyperparameters should be discarded in favour of a tailored approach to each individual dataset. These assumptions include transferring all possible layers or none, training all layers at the same learning rate or freezing some, and using L2SP regularization or L2 regularization for all layers. 
Our work demonstrates that optimal transfer learning non-binary hyperparameters are dataset dependent and strongly influenced by how well the pre-trained weights fit the target task. For a particular dataset, optimal non-binary transfer learning hyperparameters can be determined based on the difference between model performance when fixed features are used and when the full model is trained from random initialization as shown in Table \ref{predicting}. We recommend using the settings shown on the left in this table for target datasets where the difference is negative and settings shown on the right for positive differences. Target datasets for which the pre-trained weights are well suited and target datasets for which they are not result in large differences in this value. These differences should still be pronounced even if suboptimal learning hyperparameters are used for this initial test due to limited resources for hyperparameter search. This heuristic for determining optimal hyperparameters should be useful in most transfer learning for image classification cases. The normalized Fisher Ratio may be useful in some cases, however, care should be taken because the differences are not as pronounced. The EMD domain similarity measure should not be used to determine transfer learning hyperparameters.

\section{Broader Impact \label{impact}}
This research is focused on improving transfer learning for small target image classification datasets. The positive impacts are easy to observe in improved performance of models utilising transfer learning in a wide range of real world applications. For example, faster and more accurate medical diagnostics where there is limited data available. 

However, improvements to image classification models can have positive or negative impacts. More accurate models could reduce errors but result in overconfidence and more reliance on results in areas where the full limitations are not well understood and taken into account. Improved image classification models could also be used in areas with potential negative impacts, for example:
\begin{itemize}
\item use of intrusive facial recognition systems 
\item discrimination based on subtle items such as religious symbols
\end{itemize}

\section{Limitations and future work \label{limitations}}
The main aim of this work is to highlight the need for a non-binary approach to achieving optimal performance in transfer learning applied to image classification on small target datasets. The relationship between the source and the target dataset and how well the pre-trained weights fit the target task has been shown to be important to transfer learning performance numerous times \cite{ngiam2018domain,mahajan2018exploring,cui2018large,ge2017borrowing,sabatelli2018deep,li2018explicit,li2019delta,wan2019towards,ng2015deep,zoph2020rethinking,li2020rethinking}. However, the guidance as to how to set transfer learning hyperparameters based on the heuristics presented in this paper is based only on the experiments on the four datasets outlined in this paper. Care should be taken if applying this guidance to new datasets, particularly outside of image classification.

\bibliographystyle{plain}
\bibliography{main}

\end{document}